\documentclass[journal,twoside, print]{ieeecolor}
\usepackage{generic}
\usepackage{cite}
\usepackage{amsmath,amssymb,amsfonts}
\usepackage{algorithmic}
\usepackage{tikz}
\usepackage{graphicx}
\usepackage{algorithm,algorithmic}
\usepackage{hyperref}
\hypersetup{hidelinks=true}
\usepackage{textcomp}

\usepackage{mathtools}
\let\labelindent\relax
\usepackage{enumitem}
\usepackage{tikz}
\usepackage{wasysym}
\usepackage{centernot}
\usepackage{makecell}
\usepackage{bbm}
\usepackage{etoolbox}

\usetikzlibrary{automata,arrows.meta}
\newtheorem{definition}{Definition}
\newtheorem{proposition}{Proposition}
\newtheorem{problem}{Problem}

\def\BibTeX{{\rm B\kern-.05em{\sc i\kern-.025em b}\kern-.08em
    T\kern-.1667em\lower.7ex\hbox{E}\kern-.125emX}}
\begin{document}

\title{Reward Machines for Signal Temporal Logic}

\author{Alper Kamil Bozkurt, Shangtong Zhang, and Yuichi Motai
\thanks{This work was supported by the Commonwealth Cyber Initiative HV-2Q25-035, HC-2Q25-033, and the Central Virginia Node under the award VV-1Q26-001.}
\thanks{A. K. Bozkurt and Y. Motai are with the Department of Electrical \& Computer Engineering, Virginia Commonwealth University, Richmond, VA, USA (e-mail: {\tt\small \{bozkurta,ymotai\}@vcu.edu}), and S. Zhang is with the Department of Computer Science, University of Virginia, Charlottesville, VA, USA (e-mail: {\tt\small xdm2bt@virginia.edu}).}}

\maketitle

\begin{abstract}
Signal temporal logic (STL) provides a formal language for specifying real-time properties of real-valued observations, along with a quantitative robustness score for monitoring satisfaction. Control synthesis from STL specifications is of interest since manual controller design becomes infeasible as real-world systems grow in complexity. Moreover, many modern autonomous and AI-enabled systems lack accurate and complete system models, which makes optimization-based synthesis approaches unsuitable and motivates learning-based control. Prior work uses STL robustness scores as rewards in reinforcement learning (RL) to obtain control policies satisfying given specifications; however, robustness depends on execution history, leading to intractable state space expansion for general long-horizon specifications with arbitrarily nested temporal operators. This work introduces a novel automata-based approach that provides an efficient memory mechanism and associated Markovian rewards suitable for RL frameworks. Our approach constructs a timed alternating automaton from the given STL specifications, augments the state space with automaton locations and clock valuations, and derives rewards from the automaton acceptance condition. We empirically demonstrate that our approach learns policies that achieve higher robustness scores and satisfaction rates than those learned by existing approaches using robustness-based rewards.
\end{abstract}


\section{Introduction}

Signal temporal logic (STL) is a formal specification language for expressing requirements over real-time, real-valued signals \cite{maler2004monitoring}. STL combines numerical predicates, expressed as inequalities over signal values, with metric temporal operators that impose explicit real-time constraints. STL enables systematic verification by evaluating execution traces with a robustness score \cite{donze2010robust} that captures not only whether a specification is satisfied, but also how strongly it is satisfied, which is critical for noisy signals or imperfect models. These properties make STL well suited to time-critical control systems with continuous or hybrid dynamics, and it has been applied successfully in domains such as robotics \cite{gundana2021event}, traffic \cite{pigozzi2021mining}, and medical systems \cite{mambakam2024mining}. Over the past two decades, STL has been extended in many directions, including online monitoring methods for robustness \cite{donze2013efficient, deshmukh2017robust, mehdipour2024generalized} and richer robustness notions that account not only for spatial perturbations, but also for timing perturbations and other forms of uncertainty \cite{rodionova2022temporal}. Despite the practical value of runtime verification for existing control systems, synthesizing controllers directly from STL specifications is required since manual controller design is impractical for many real-world systems \cite{seshia2015combining}.

Control synthesis from STL specifications via optimization for the systems with available models has been widely studied (e.g., \cite{raman2014model, raman2015reactive, lindemann2021reactive, charitidou2022receding, yao2023multitask}). However, as modern autonomous systems grow more complex and incorporate more AI components, high-fidelity models suitable for standard optimization techniques are often unavailable, necessitating data-driven learning. As a result, a growing body of work has sought to integrate STL directly into learning-based control, leveraging its quantitative robustness scores as rewards in reinforcement learning (RL) pipelines. However, the history-dependent semantics of robustness scores over traces violates the Markov property, a common assumption in standard RL frameworks. Existing approaches address this issue either by augmenting the state with previously visited states \cite{aksaray2016q, muniraj2018enforcing}, which is intractable for long-horizon specifications; or by restricting attention to limited fragments of STL \cite{venkataraman2020tractable, kalagarla2021model, wang2024tractable, varnai2019prescribed, liu2023learning}. To the best of our knowledge, there is no existing RL approach that considers full STL while remaining more tractable than state augmentation with the entire history.

In this work, we mitigate the history dependence of STL satisfaction by constructing reward machines (RMs), which provide an efficient memory mechanism and induce Markovian rewards, thereby enabling RL-based control synthesis from STL specifications. Our contributions are as follows:
\begin{itemize}[leftmargin=*]
    \item We introduce a novel automata-based framework for learning controllers from STL specifications. We model stochastic control systems as semi-Markov decision processes (SMDPs) and adopt event-based STL semantics, which enables the derivation of one-clock alternating timed automata (OCATAs) \cite{brihaye2017mightyl} from the specifications.
    \item We construct STL-RMs from the derived OCATAs using their acceptance conditions, while additionally incorporating robustness to observation perturbations inspired by the differentiable rewards from \cite{bozkurt2026accelerated}. Beyond providing rewards, our RMs maintain a list of automaton locations with clock valuations that serves as memory for state augmentation, which makes the rewards Markovian and compatible with off-the-shelf RL algorithms. We formalize that any control policy learned using our RMs that achieves the maximum cumulative reward of $1$ satisfies the given STL specification with probability~$1$.
    \item We show that our approach outperforms existing methods by learning control policies faster and achieving higher satisfaction rates on long-horizon specifications across several simulated experiments.
\end{itemize}

The rest of the paper is organized as follows. In Section \ref{sec:related_work}, we review the related work, and in Section \ref{sec:prelim}, we provide necessary background information and establish our notation. We introduce our approach in Section \ref{sec:rms_for_stl} and present our experimental results in Section \ref{sec:exp}. Finally, we draw conclusions in Section \ref{sec:conclusion}.

\section{Related Work}\label{sec:related_work}

Prior work on controller synthesis from STL specifications falls into two categories, depending on whether a system model is assumed to be available: model-based and model-free. We discuss prominent approaches and their drawbacks in both of these categories below. We refer to \cite{yin2024formal} for detailed discussions.

\subsection{Model-Based Synthesis Approaches}
Previous research has largely focused on establishing mixed-integer programs (MIPs) for synthesizing controllers from STL specifications \cite{liu2017communication, kress2018synthesis, belta2019formal, kurtz2022mixed}. A common approach is to utilize model predictive control (MPC), where, at each time step, an optimal control policy over a finite horizon is obtained by the MIP formulated based on the system dynamics; this procedure is then repeated iteratively in a receding-horizon manner \cite{raman2014model}. Such approaches have been extended to worst-case scenarios \cite{yu2026signal}, adversarial settings \cite{raman2015reactive}, systems under disturbance \cite{zhang2025decomposition}, uncertain or stochastic environments \cite{sadigh2016safe, jha2018safe, farahani2018shrinking}, resilient control \cite{chen2023stl}, multiple objectives \cite{yao2025model}, and unbounded specifications \cite{ilyes2025receding}.

A main issue in these MPC formalisms is that shorter planning horizons can lead to undesirable, myopic solutions, whereas longer horizons can be computationally expensive. Some approaches propose using control barrier functions (CBFs) for computational efficiency; however, they typically consider only fragments of STL \cite{lindemann2018control, gundana2021event}, assume linearity \cite{yang2020continuous}, or require additional reachable set computation \cite{yu2024continuous}. Another line of studies, e.g., \cite{pant2017smooth, lindemann2019robust, haghighi2019control, mehdipour2019arithmetic, gilpin2020smooth, takayama2025stlccp}, proposes smoothed versions of robustness to enable gradient-based optimization for faster computation, and has also explored combining these methods with neural networks \cite{leung2023backpropagation, meng2023signal} via backpropagation. Others include tube-based \cite{vlahakis2024probabilistic, das2025approximation}, prescribed performance control (PPC) \cite{lindemann2017prescribed, chen2024cooperative, liu2025controller}, time-interval decomposition \cite{yu2023model, yang2024signal}, system transformation \cite{lai2025continuous}, all introducing additional requirements, e.g., on STL formulas, or system dynamics.

Overall, model-based synthesis for STL has been an active research area, yielding many studies. However, the history dependence of STL robustness scores remains a key obstacle in control synthesis. This dependence increases the computational burden in MILP formalisms for longer planning horizons and can cause vanishing/exploding gradient issues when backpropagating through long histories. Additionally, all these approaches rely on the assumption that a system model is available, limiting their applicability.

\subsection{Model-Free Learning Approaches}

Modern RL has achieved strong empirical performance in learning reward-maximizing controllers directly from interaction data, without requiring an explicit dynamics model \cite{recht2019tour}. This success has motivated the use of RL for control synthesis from STL specifications, by employing robustness scores as the rewards in RL objective \cite{balakrishnan2019structured, varnai2020robustness, hamilton2022training}. A central challenge is that the satisfaction rates and robustness scores are calculated over the entirety of traces, making them history-dependent, thereby breaking the Markov property assumed by most RL formulations. This non-Markovian dependence can destabilize learning and may lead to poor performance, or even divergence, particularly for value-based methods such as Q-learning and actor-critic algorithms.

A common technique to restore Markovian structure is augmenting the state space with the recent history of the visited states, where the required history length is determined by the temporal structure of the STL specification \cite{aksaray2016q, muniraj2018enforcing, ikemoto2022deep2, wang2024synthesis}. While conceptually simple, this approach can dramatically increase the state dimension, and the resulting complexity becomes prohibitive for long-horizon specifications. To mitigate this blow-up, several works restrict attention to tractable fragments of STL or introduce alternative intermediates that avoid full-history augmentation. Examples include augmenting the state with compact bookkeeping variables for limited nesting \cite{venkataraman2020tractable, kalagarla2021model, wang2024tractable}, prescribed performance control formulations \cite{varnai2019prescribed}, sampling-based planning methods \cite{tian2022two}, learning with control barrier functions \cite{liu2023learning}, and funnel-based control \cite{saxena2023funnel}. Despite these advances, there remains no model-free approach that scales to full STL while avoiding intractable history-based state augmentation.

A line of work closely related to ours focuses on crafting automata-based rewards for RL by compiling temporal-logic specifications into automata and learning over the product system. Most existing methods build rewards from omega automata obtained from logics without real-time constraints such as linear temporal logic (LTL), e.g., \cite{hahn2019omega, hasanbeig2019reinforcement, bozkurt2020control, cai2020learning, wen2021probably, oura2024bounded, bozkurt2024learning, kantaros2024sample, cai2025safety}, and a few consider reward shaping for a given timed automaton \cite{dole2021event}, a formalism extending transition systems with clock variables and time constraints. To the best of our knowledge, however, prior work has not constructed rewards from an automaton directly derived from STL in a way that is compatible with standard RL settings. While STL can be translated into continuous-time signal transducers \cite{lindemann2021reactive}, those representations are not well matched to RL, where the agent typically receives point-wise state observations at discrete decision times. In this work, we translate STL specifications into OCATAs \cite{brihaye2017mightyl} and design rewards to accommodate their conjunctive branching structure and the associated acceptance condition, which can be incorporated into RL via augmentation with automaton states and clock variables.

\section{Preliminaries}\label{sec:prelim}

\subsection{Semi-Markov Decision Processes}
We model the interaction between controllers and the environments as SMDPs, which extend standard MDPs by allowing stochastic, real-valued durations between successive observations.

\begin{definition}[SMDP]
An SMDP is a tuple $\mathcal{M} = (S, s_0, A, P)$, where $S$ is a set of states, $s_0$ is the initial state, $A$ is a set of actions\footnote{We primarily consider continuous state and action spaces, although our approach also applies to discrete settings.}, and $P$ is a probabilistic transition function. 
We write $P(s',\tau' \mid s,a)$ for the probability density (or probability mass, in the discrete case) of transitioning from the state $s\in S$ to the state $s'\in S$ when taking action $a \in A$, after a nonnegative elapsed time $\tau'\in\mathbb{R}_{\ge 0}$.
\end{definition}

The time value $\tau$ captures the duration between observing $s$ and observing $s'$, including both transition and dwelling time; this timing information is explicit in SMDPs, whereas it is abstracted away in discrete-time MDPs.

A control policy $\pi: (S\times \mathbb{R}_{\geq 0})^{+} \mapsto A$ for an SMDP $\mathcal{M}$ is a mapping that selects an action based on the history of states with duration stamps. Starting from the initial pair\footnote{For simplicity, we assume a fixed initial state $s_0$ and duration $\tau_0=0$. Our approach extends directly to random initial states and duration.} $(s_0,\tau_0)$, execution of a policy $\pi$ in an SMDP $\mathcal{M}$ generates a \emph{timed path}, an infinite sequence of duration-stamped states $\sigma \coloneqq (s_0,\tau_0)(s_1,\tau_1)\dots$ according to the induced Markov chain (MC) $\mathcal{M}_\pi$, where the increment $\tau_t$ is the duration between observations of $s_t$ and $s_{t+1}$ for all $t>0$. We use $\sigma[t]$, $\sigma[{:}t]$, and $\sigma[t{:}]$ to denote the timed state $(s_t,\tau_t)$, the prefix $(s_0,\tau_0)\dots(s_t,\tau_t)$, and the suffix $(s_t,\tau_t)(s_{t+1},\tau_{t+1})\dots$, respectively.

For a given bounded reward function\footnote{We consider state-based reward functions for simplicity; our approach is compatible with state-action-based reward functions as well.} $R: (S\times \mathbb{R}_{\geq 0})^{+} \mapsto \mathbb{R}$ mapping timed path prefixes to real-valued scalars, and a given discount factor $\gamma \in [0,1)$, the return $G(\sigma)$ of a path $\sigma$ is the sum of discounted rewards $G(\sigma) = \sum_{t=0}^\infty \gamma^tR(\sigma[{:}t])$. In standard formulations, discounting uses a fixed factor $\gamma$; but this can be relaxed to allow for state-dependent discounting as described in \cite{bozkurt2020control}.

The objective in an SMDP $\mathcal{M}$ with a reward function $R$ is to find an optimal policy that maximizes expected return $\pi^* = \mathrm{argmax}_\pi \mathbb{E}_{\sigma \sim \mathcal{M}_\pi} \left[G(\sigma)\right]$. When rewards are Markovian (i.e., the reward function $R:S\times \mathbb{R}_{\geq 0} \mapsto\mathbb{R}$ maps based only on the last observed state and duration), it is sufficient to consider memoryless policies $\pi:S\mapsto A$, and an optimal policy can be learned using off-the-shelf RL tools when the transition function $P$ is unknown. However, in history-dependent settings (e.g., STL objectives), applying RL typically requires a Markovian reward design, which can be achieved by augmenting the state space with a tractable set of memory variables.

\subsection{Event-Based Signal Temporal Logic}
STL provides a language to specify temporal properties of real-valued observations using predicates, Boolean connectives, and temporal operators \cite{maler2004monitoring}.

\begin{definition}
An STL predicate is an inequality $\mu \geq 0$, where $\mu:S \to \mathbb{R}$ is a function mapping a state to a scalar. STL specifications can be formulated recursively by the following grammar:
\begin{equation}
    \varphi \coloneqq  \mu \geq 0 \ \mid\  \neg \varphi \ \mid\  \varphi_1 \wedge \varphi_2 \ \mid\  \varphi_1 \,\mathrm{U}_I\, \varphi_2 \ , \quad \mu \in \Lambda\,,    \label{eq:stl}
\end{equation}
where $\neg$ is negation, $\wedge$ is conjunction, $\mathrm{U}$ is the until operator, and $I \subseteq \mathbb{R}_{\geq 0}$ is a time interval (closed, open, or half-open) with nonnegative rational or infinite endpoints ($\mathbb{Q}_{\geq0} \cup \{\infty\}$), and $\Lambda$ is the set of predicate functions.
\end{definition}

We also use the standard derived operators: disjunction $\varphi_1 \vee \varphi_2 := \neg(\neg \varphi_1 \land \neg \varphi_2)$, implication $\varphi_1 \implies \varphi_2 := \neg \varphi_1 \lor \varphi_2$, true $\top \coloneqq \varphi \vee \neg \varphi$, finally $F_I \varphi := \top \ \mathrm{U}_I \varphi$, and globally $G_I \varphi := \neg F_I \neg \varphi$. When $I=[0,\infty)$, we omit the interval subscript from temporal operators.

We adopt an \emph{event-based} semantics defined over infinite timed paths $\sigma=(s_0,\tau_0)(s_1,\tau_1)\dots$ (rather than continuous-time signal semantics) to align with standard RL formalisms. We write $\sigma[t{:}] \models \varphi$ to denote that the timed suffix $\sigma[t{:}]$ satisfies the specification $\varphi$. Following the formulation in \cite{ouaknine2007decidability} for metric interval temporal logic (MITL), we define the semantics recursively by
\begin{align}
& \sigma[t{:}] \models \mu \geq 0 && \iff \mu(s_t) \geq 0 \notag \\
& \sigma[t{:}] \models \neg \varphi && \iff \sigma[t{:}] \not\models \varphi \notag \\
& \sigma[t{:}] \models \varphi_1 \wedge \varphi_2 && \iff \sigma[t{:}] \models \varphi_1 \text{ and } \sigma[t{:}] \models \varphi_2 \notag \\
& \sigma[t{:}] \models \varphi_1 \mathrel{\mathcal{U}_I} \varphi_2 && \iff \exists i \text{ s.t. } \tau_i - \tau_t \in I, \ \sigma[i{:}] \models \varphi_2, \notag \\
& && \hspace{3.7em} \text{ and } \forall t \leq j < i, \  \sigma[j{:}] \models \varphi_1. \notag
\end{align}
Similarly, we define the semantics of spatial robustness scores as follows: 
\begin{align}
& \rho(\sigma[t{:}],\ \mu \geq 0) = \mu(s_t) \notag \\
& \rho(\sigma[t{:}] \ \neg \varphi) =  - \rho(\sigma[t{:}] \ \varphi) \notag \\
& \rho(\sigma[t{:}],\ \varphi_1 \wedge \varphi_2) = \max\{\rho(\sigma[t{:}],\ \varphi_1), \ \rho(\sigma[t{:}],\ \varphi_2)\} \notag \\
& \rho(\sigma[t{:}], \ \varphi_1 \mathrel{\mathcal{U}_I} \varphi_2) = \sup_{i\in \{k | \tau_k - \tau_t \in I\}}  \notag \\
& \hspace{7em} \min\left\{\rho(\sigma[i{:}],\ \varphi_2), \ \inf_{t \leq j < i } \rho(\sigma[j{:}]\ \varphi_1)\right\}. \notag
\end{align}
The inherent history-dependence due to $\mathrm{inf}$ and $\mathrm{sup}$ operations applied over path fragments makes the spatial robustness scores unsuitable for serving as Markovian rewards in RL for long-horizon specifications. Thus, we do not utilize these robustness scores directly in our approach; instead, we employ them for evaluation and comparison.

\section{Problem Formulation}

We study the efficient learning of control policies that satisfy given STL specifications in unknown environments. We adopt SMDPs as the system formalism. Unlike standard discrete-time MDPs, in which time is abstracted as a sequence of integer-valued steps, and continuous-time MDPs, in which states are observed and actions are selected continuously, SMDPs explicitly represent the elapsed time between consecutive observations. This representation enables real-time constraints to be expressed through event-based STL while retaining compatibility with off-the-shelf RL algorithms that operate on pointwise observations.

A key challenge in this setting is accounting for stochastic transitions. Repeated executions of the same policy may produce different successor states and transition durations, thereby generating different timed paths. Our primary objective is therefore to learn a policy that maximizes the satisfaction probability, defined as the probability that a timed path generated under the policy satisfies the STL specification. As a secondary objective, we incorporate robustness into this probability-maximization framework by modeling probabilistic satisfaction of predicates under supposed random perturbations. We formalize the resulting joint problem as follows.

\begin{problem}\label{problem}
Given an SMDP $\mathcal{M}$ with an unknown transition function $P$ and an STL specification $\varphi$, design a mechanism $\mathcal{R}_\varphi$ that generates memory states and corresponding Markovian rewards, thereby enabling efficient RL to obtain an optimal policy $\pi^*$ that maximizes the probability of satisfying $\varphi$:
\begin{align} 
    \pi^* \coloneqq& \mathrm{argmax}_\pi \Pr\nolimits_{\sigma\sim\mathcal{M}_\pi}(\sigma\models\varphi) 
\end{align}
Here, $\Pr(\cdot)$ accounts for both (i) the stochasticity of a timed path $\sigma$ generated by the SMDP $\mathcal{M}$ under a policy $\pi$ and (ii) the probabilistic satisfaction of the specification $\varphi$ by $\sigma$ due to modeled random perturbations of its predicates.
\end{problem}

Since we consider stochastic environments, the proposed formulation naturally focuses on maximizing satisfaction probability. We do not optimize the expected robustness degree, as doing so may favor policies that achieve high robustness on a small subset of trajectories while frequently violating the specification. Instead, we use expected robustness only as an additional evaluation metric alongside satisfaction probability. Nevertheless, our probabilistic interpretation of predicate satisfaction provides a principled and tunable means of incorporating robustness into the probability-maximization objective.

\begin{figure*}
\centering
\resizebox{0.9\textwidth}{!}{
    \begin{tikzpicture}[>=Stealth,->,semithick,
        every state/.style={draw,circle,minimum size=8mm,inner sep=0pt},
        small/.style={draw,circle,minimum size=5mm,inner sep=0pt}
    ]
        \node[state,initial] (l0) at (0,0) {$l_0$};
        \node[small] (or) at (2,0) {$\vee$};
        \node[state,accepting] (l1) at (4,0) {$l_1$};
        \node[small] (and) at (7,0) {$\wedge$};
        \node[state] (l2) at (9, 0) {$l_2$};
        \node[state,accepting] (l3) at (12.5, 0) {$l_3$};
        \node[state] (l4) at (12.5, 1.5) {$l_4$};

        \path (l0) edge node[below] {\makecell{$\alpha{=}\top$ \\ $\kappa{=}\top$}} (or);
        \path (or) edge[bend right=30] node[above] {$\centernot\cent{=}\bot$} (l0);
        \path (or) edge node[above] {$\centernot\cent{=}\bot$} (l1);

        \path (l1) edge[out=120,in=60,looseness=6] node[above] {\makecell{$\alpha{=}``|x{-}x_\textit{f}|{\leq}1"$ \\ $\kappa{=}\top$ \\ $\centernot\cent{=}\bot$}} (l1);
        \path (l1) edge node[below] {\makecell{$\alpha{=}``|x{-}x_\textit{f}|{>}1"$ \\ $\kappa{=}\top$}} (and);
        \path (and) edge[bend right=30] node[above] {$\centernot\cent{=}\bot$} (l1);
        \path (and) edge node[above] {$\centernot\cent{=}\top$} (l2);
        \path (and) edge node[below] {reset} (l2);

        \path (l2) edge[out=120,in=60,looseness=6] node[above] {\makecell{$\alpha{=}``|x{-}x_\textit{f}|{>}1"$ \\ $\kappa{=}``\cent {\leq} 1"$ \\ $\centernot\cent{=}\bot$}} (l2);
        \path (l2) edge node[below] {\makecell{$\alpha{=}``|x{-}x_\textit{f}|{\leq}1"$ \\ $\kappa{=}``\cent {\leq} 1"$ \\ $\centernot\cent{=}\bot$}} (l3);
        \path (l2) edge node[above] {\makecell{$\alpha{=}\top$ \\ $\kappa{=}``\cent {>} 1"$ \\ $\centernot\cent{=}\bot$}} (l4);

        \path (l3) edge[out=30,in=330,looseness=3] node[right] {\makecell{$\alpha{=}\top$ \\ $\kappa{=}\top$}}  (l3);

        \path (l4) edge[out=30,in=330,looseness=3] node[right] {\makecell{$\alpha{=}\top$ \\ $\kappa{=}\top$}}  (l4);
    \end{tikzpicture}
}
\caption{Illustration of an OCATA derived from the STL formula $\varphi_f=FG\big(\neg``\mu_f{\geq}0" \implies F_{[0,1]}``\mu_f{\geq}0"\big)$. Here, $\mu_f(s)\coloneqq 1-|x{-}x_\textit{f}|$ is a predicate function measuring how close the position component $x$ of a state $s$ is to the target position $x_\textit{f}$. Larger circles $(L=\{l_0,l_1,l_2,l_3,l_4\})$ represent automaton locations with accepting locations shown as double circles $(\{l_1,l_3\})$. Arrows represent transitions; and smaller circles, $\vee$ and $\wedge$, indicate disjunctive (nondeterministic/existential) and conjunctive (universal) branching, respectively. Symbols $\alpha$, $\kappa$, $\centernot\cent$, and $\cent$ denote input letters, clock constraints, clock-reset flags, and the clock variable, respectively.
}
\label{fig:ocata}
\end{figure*}
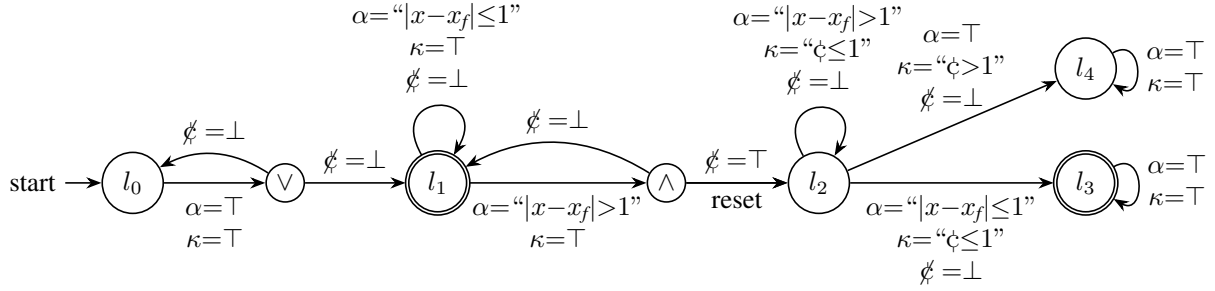

\section{Reward Machines for Signal Temporal Logic}\label{sec:rms_for_stl}

We address Problem~\ref{problem} by introducing extended RMs constructed for OCATAs derived from the STL specifications. Our RMs extend the standard definition \cite{icarte2022reward} by allowing nondeterministic, conjunctive, and probabilistic transitions, together with the B\"uchi acceptance condition, in order to capture the event-based semantics of full STL under observation perturbations. We then provide a procedure for composing the product of an SMDP and an RM, yielding an augmented model with Markovian rewards. Finally, we formalize that any policy achieving the maximum expected return of 1 under these rewards is guaranteed to satisfy the STL specification with probability 1. We describe each step in detail in the following subsections.

\subsection{One-Clock Alternating Timed Automata}

Our procedure begins by translating a given specification to an OCATA with a B\"uchi acceptance condition (repeated reachability) as in \cite{brihaye2017mightyl}\footnote{Although this construction is presented for MITL, it applies directly to STL as well.}. OCATAs are timed automata with a single clock variable that allow for disjunctive and conjunctive transitions. We first introduce the grammar of clock constraints used in the OCATA construction. Let $K$ be the finite set of clock-constraint formulas generated by
\begin{align}
\kappa \coloneqq \ \cent <c \ \mid \ \cent \leq c \ \mid \  \neg \kappa  \ \mid \ \kappa_1 \wedge \kappa_2 \label{eq:clock_constraints}
\end{align}
where $\cent$ denotes the single clock variable, and $c \in \mathbb{Q}_{\geq0}$ is an arbitrary nonnegative rational number. Other relations ($\{>,\geq,=,\neq\}$) and logical operators ($\vee,\implies,\top$) can be derived as usual. 

In a deterministic one-clock timed automaton, receiving an input $\alpha \in \Sigma$ under a satisfied clock constraint $\kappa \in K$ triggers a unique transition from a current location $l \in L$ to a destination $l' \in L$ where $\Sigma$ is a finite alphabet and $L$ is a finite set of locations. Each transition is additionally labeled with a reset flag $\centernot\cent \ \in \{\top,\bot\}$ indicating whether the clock $x$ is reset with the transition. In an OCATA, however, a $(l,\alpha,\kappa)$ triple may induce not only a single transition but also disjunctive ($\vee$) and conjunctive ($\wedge$) combinations of transitions. To represent such branching behavior, we define a finite set of destination formulas $D$ using the grammar
\begin{align}
d \coloneqq \ (l,\centernot\cent) \ \mid \ d_1 \wedge d_2 \ \mid \ d_1 \vee d_2, \label{eq:destination_formulas}
\end{align}
where $l\in L$ is a destination location and $\centernot\cent \ \in \{\top,\bot\}$ is the clock-reset flag. We now provide the definition of an OCATA:

\begin{definition}
An OCATA is a tuple $\mathcal{A}=(L,l_0,\Sigma,K,D,\allowbreak\delta, F)$, where $L$ is a set of locations, $l_0 \in L$ is the initial location, $\Sigma=2^{\Lambda}$ is an alphabet where $\Lambda$ is a predicate set from \eqref{eq:stl}, $K$ is a set of clock constraints defined by \eqref{eq:clock_constraints}, $D$ is a set of destination formulas defined by \eqref{eq:destination_formulas}, $\delta : L \times \Sigma \times K \to D$ is a total transition function, and $F \subseteq L$ is a set of accepting locations.
\end{definition}

An OCATA starts in the initial location $l_0$ with a clock value $x=0$, and makes transitions as observations from the SMDP are received. An OCATA state is determined by a pair $q=(l,v)$, where $l\in L$ is the current location and $v\in\mathbb{R}_{\geq 0}$ is the current clock valuation. After observing $\sigma[t]=(s_t,\tau_t)$ at time step $t$ in a location $l$, the OCATA advances the clock value by $\tau_t$, computes the set of satisfied predicates $\alpha_t=\{\mu\in\Lambda \mid \mu(s_t)>0\} \in \Sigma$, and selects the clock constraint $\kappa_v \in K$ that is satisfied under the updated valuation $v$. The OCATA then makes transitions according to the destination formula $d=\delta(l,\alpha,\kappa)$, where the nature of the transitions is determined by the $\vee$ and $\wedge$ operators in the formula $d$. The disjunction operator $\vee$ induces nondeterministic transitions (existential choice) where at least one transition should lead to acceptance. In contrast, the conjunctive operator $\wedge$ spawns new threads of computations by creating different copies of OCATA states (universal choice) where each copy must lead to acceptance.

The semantics of transitions are formally defined based on acceptance as follows. An OCATA $\mathcal{A}$ in an automaton state $(l,v)$ accepts the timed suffix $\sigma[t{:}]$ with respect to a destination formula $d$ if and only if (iff) one of the following recursively holds:
\begin{itemize}[leftmargin=1em]
    \item $d = d_1 \vee d_2$ and $\mathcal{A}$ in $(l,v)$ accepts $\sigma[t{:}]$ for $d_1$ or $d_2$;
    \item $d = d_1 \wedge d_2$ and $\mathcal{A}$ in $(l,v)$ accepts $\sigma[t{:}]$ for $d_1$ and $d_2$;
    \item $d=(l',\centernot\cent)$ and $\mathcal{A}$ in $(l',v')$ with $v'=[\neg\hspace{-0.2em}\centernot\cent\hspace{0.1em}]v+(\tau_{t+1}-\tau_t)$ accepts $\sigma[t{+}1{:}]$ with respect to $d'=\delta(l',\alpha_{t+1},\kappa_{v'})$;
\end{itemize}
where $[\neg\hspace{-0.2em}\centernot\cent\hspace{0.1em}]$ is $0$ if $\centernot\cent \ =\top$ else $1$. A run of an OCATA induced by a timed path is a computation tree, due to the threads spawned by conjunctive transitions. Moreover, disjunctive transitions introduce nondeterminism, yielding multiple runs from which the OCATA may choose. A run is accepting iff every infinite computational path visits some accepting location in $F$ infinitely often (the B\"uchi condition). The OCATA accepts a timed path iff there exists an accepting run induced by that timed path.

Fig.~\ref{fig:ocata} illustrates an OCATA derived from an STL formula. The formula intuitively requires that the target region, defined as the set of positions within distance $1$ of the target $x_\textit{f}$, is eventually reached and, thereafter, whenever this region is left, it must be returned to within $1$ time unit.
The disjunction $\vee$ introduces a nondeterministic choice between transitions to $l_0$ and $l_1$. Transitioning to $l_1$ corresponds to guessing that the target region has been reached and that any future deviation can be corrected within the required time bound; transitioning to $l_0$ represents the opposite guess. When an observation indicates that the region is left, the conjunction $\wedge$ spawns two branches to $l_1$ and $l_2$. The $l_2$ branch enforces the $1$-time-unit return condition by setting a timer via resetting the clock: if the region is re-entered in time, the OCATA moves to the accepting sink $l_3$; otherwise, it moves to the rejecting sink $l_4$. The $l_1$ branch keeps the OCATA in $l_1$ to handle potential future departures, and its induced computation path is accepting as $l_1$ is an accepting location. However, because $\wedge$ requires universal acceptance, all computation paths generated by the $l_2$ branch must also be accepting, ensuring that the region is always returned to within $1$ time unit. Next, we explain how to construct an RM that reflects this OCATA acceptance condition.

\subsection{Reward Machine Construction}
We construct RMs for given STL specifications based on the OCATAs derived from the specifications. The main challenge is to design rewards that (i) encode the B\"uchi acceptance condition requiring some accepting locations to be visited infinitely often, (ii) handle the disjunctive and conjunctive transitions, and (iii) incorporate robustness to state perturbations.

We adopt the state-based rewarding and discounting scheme of \cite{bozkurt2020control}. The idea is to provide a reward of $r\in(0,1]$ whenever an accepting location $l \in F$ is visited, and a reward of $0$ for visiting other locations $l\in L\setminus F$. The future rewards are discounted with factors of $\gamma=1-r$ in accepting and $\gamma'=1-r^2$ in non-accepting locations, reflecting the fact that visiting non-accepting locations is irrelevant to satisfying the B\"uchi condition. Maximizing the expected return under these rewards and discount factors corresponds to maximizing the probability of satisfying the B\"uchi condition \cite{bozkurt2020control}.

The nondeterminism in OCATAs due to the disjunctive transitions can be encoded as $\epsilon$-actions in RMs that the RL agent can select during learning. In such formulations, the expected return establishes a lower bound on the satisfaction probability in general; and if the OCATA is limit-deterministic in a suitable way, then the lower bound becomes equality \cite{hahn2019omega}. To the best of our knowledge, it is not known whether STL (or MITL) formulas can always be translated into OCATAs that are limit-deterministic in this way. Accordingly, we focus on the lower-bound guarantee rather than equivalence.

In order to capture the conjunctive transitions, we keep track of the automaton states for each copy of the OCATA created by conjunctions in a memory list, and the average value of the rewards associated with each copy is provided as a joint reward. This ensures that the maximization of the reward in each copy maximizes the joint return. In order to establish a tractable probabilistic framework, we extend this formalism by associating each copy with a probability. Specifically whenever two copies are spawned, they split the probability associated with their parent. This allows us to keep a single copy for each clock valuation; the conjunctive splits are just going to affect the probabilities over locations if they do not reset the clock; only clock resets will trigger a copy.

To handle conjunctive transitions, we track the automaton state for each OCATA copy spawned by conjunctions using a memory list. We then provide a joint reward given by the average of the rewards across these copies. This construction ensures that maximizing the return in each copy contributes to maximizing the joint return. To obtain a tractable probabilistic representation, we associate each copy with a probability mass: whenever a conjunction spawns multiple copies, the parent probability is split among the children. The conjunctive branching that does not reset the clock only redistributes probability mass over locations; in contrast, clock resets trigger the creation of a new copy (i.e., a new memory entry). This lets us maintain a single copy per clock valuation.

Finally, we incorporate robustness by introducing pseudo-perturbations in evaluations of STL predicates. Concretely, we treat the predicate function output $\mu(s)$ as if it were observed through additive noise, e.g., drawn from a normal or logistic distribution, inducing probabilities over predicates. Following the idea of differentiable rewards for LTL in \cite{bozkurt2026accelerated}, we define the probability of satisfying a predicate for an observed state $s$ as $\textsc{Pr}(\mu>0)=h(\mu(s))$ where $h$ is a cumulative distribution function (CDF) (e.g., the sigmoid for logistic noise). The probability of observing each input letter $\alpha\in\Sigma=2^\Lambda$ is then calculated as the product of $h(\mu(s))$ for $\mu \in \alpha$ and $1-h(\mu(s))$ for $\mu \not\in \alpha$, which corresponds to the probability of making a transition with $\alpha$. We now formally define the STL-RM:

\begin{definition}
An STL-RM is a tuple $\mathcal{R}_\varphi=(Q, q_0, \Delta, R)$ constructed for an OCATA $\mathcal{A}_\varphi=(L,l_0,\Sigma,K,D,\allowbreak\delta, F)$ derived from an STL specification $\varphi$; $Q=([0,1]^{|L|}\times \mathbb{R}_{\geq0})^{+}$ is a memory list tracking location probabilities and clock valuations for each OCATA copy; $q_0$ is the initial memory list; $\Delta:Q\times S \times \mathbb{R}_{\geq 0} \times \mathcal{E}\mapsto Q$ is the transition function that updates the memory list where $\mathcal{E}$ is the set of $\epsilon$-actions corresponding to nondeterministic transitions in $\delta$; and $R:Q\mapsto[0,1]$ is a reward function.
\end{definition}

The transition function $\Delta$ modifies a memory list $q \in Q$ as follows.
Let $\mathbf{q}_t=(v_t^{(1)}, \mathbf{p}_t^{(1)})(v_t^{(2)}, \mathbf{p}_t^{(2)})\dots(v_t^{(n)}, \mathbf{p}_t^{(n)})$ be the current memory list where $\mathbf{p}_t^{(i)}=\{p_t^{(i,l)}\}_{l\in L}$ denotes the probability vector such that $p_t^{(i,l)}$ is the probability of being in location $l$; and $v_t^i$ denotes the clock valuation in the $i$-th copy at time step $t$. The RM $\mathcal{R}_\varphi$ starts with the initial memory list $\mathbf{q}_0=(v_0^{(1)},\mathbf{p}_0^{(1)})$ where $v_0^{(1)}=0$ and $p_0^{(1,l)}$ is $1$ for $l=l_0$ and $0$ for all $l\neq l_0$. After each observation $(s_t,\tau_t)$ from the SMDP along with an $\epsilon$-action $\varepsilon_t$ provided by the RL agent, $\mathcal{R}_\varphi$ advances each clock valuation $v_t^{(i)}$ by $\tau_t$ and makes a transition for each automaton state in the memory, i.e., $(l,v_t^{(i)})$ for every $l\in L$ with $p_t^{(i,l)}>0$ in each copy $i=1,2,\dots,n$, based on $s_t$ and $\varepsilon_t$. 

The transitions are performed by handling probabilistic inputs, nondeterministic and conjunctive branching.
Specifically,  for the observation $s_t$, the transition probabilities $p^{s_t}_\alpha$ associated with each input $\alpha$ are first computed as follows
\begin{align}
    p^{s_t}_\alpha \coloneqq& \prod_{\mu \in \alpha} \Pr(\mu\geq0) \prod_{\mu \notin \alpha} \Pr(\mu<0) \notag \\
    =&\prod_{\mu \in \alpha} h(\mu(s_t)) \prod_{\mu \notin \alpha} \left(1-h(\mu(s_t))\right).
\end{align}
The nondeterminism ($\vee$) is then resolved by choosing the branches corresponding to the $\epsilon$-action $\varepsilon_t$ provided by the RL agent. Lastly, for the conjunctions ($\wedge$), the probability of a source location $l$ is divided equally among the destinations. For example, if there are two conjunctive destinations $l'$ and $l''$ without clock resets, the probability that flows from $l$ to $l'$ (or to $l''$) is $p_t^{(i,l)}p^{s_t}_\alpha/2$. However, if one of the transitions, say $l'$, resets the clock, then the probability $p_t^{(i,l)}p^{s_t}_\alpha/2$ flows to a new entry added to the memory with a clock valuation of 0. We note that the sum of all of the probabilities always remains equal to $1$ due to the total transition function with respect to input letter, resolved nondeterminism, and equal probability splitting in conjunctions. Lastly, $R:Q\mapsto[0,1]$ is a reward function that maps a memory list $q_t$ to a reward $r_t = r\sum_i\sum_{l\in F} p^l_F$.

\begin{table}[t]
\centering
\caption{An Execution of STL-RM for OCATA from Fig.~\ref{fig:ocata}}
\label{tab:rm_execution}
\setlength{\tabcolsep}{0.2em}
\renewcommand{\arraystretch}{1.5}
\resizebox{\columnwidth}{!}{
\begin{tabular}{c | c | c | l | c}
\makecell{\textbf{Time} \\ $t$} & \makecell{\textbf{Observation} \\ $(s_t,\tau_t)$} & \makecell{$\mathcal{E}$\textbf{-Action} \\ $\varepsilon_t$ } & \makecell{\textbf{Memory List} \\ $(v_{t+1}, \mathbf{p}_{t+1})^+$} & \makecell{\textbf{Reward} \\ $r_{t+1}$} \\
\Xhline{1.2pt}
$1$ & $(\langle1.4\rangle,0.7)$ & $1$ & $1{:}\ (1.1,\langle0.0,1.0,0.0,0.0,0.0\rangle)$ & $0.1$ \\ \hline 
$2$ & $(\langle3.1\rangle,0.6)$ & $0$ & \makecell{\vspace{-0.9em}\\\hspace{-0.3em}$1{:}\ (1.7,\langle0.0,0.8,0.0,0.0,0.0\rangle)$ \\ $2{:}\ (0.0,\langle0.0,0.0,0.2,0.0,0.0\rangle)$ \vspace{0.2em}} & $0.08$ \\ \hline
$3$ & $(\langle5.0\rangle,0.9)$ & $0$ & \makecell{\vspace{-0.9em}\\\hspace{-0.3em}$1{:}\ (2.6,\langle0.0,0.6,0.0,0.0,0.0\rangle)$ \\ \hspace{-0.3em}$2{:}\ (0.9,\langle0.0,0.0,0.1,0.1,0.0\rangle)$ \\ $3{:}\ (0.0,\langle0.0,0.0,0.2,0.0,0.0\rangle)$ \vspace{0.2em}} & $0.07$ \\ \hline
$4$ & $(\langle3.8\rangle,0.3)$ & $0$ & \makecell{\vspace{-0.9em}\\\hspace{-0.3em}$1{:}\ (2.9,\langle0.0,0.6,0.0,0.0,0.0\rangle)$ \\ \hspace{-0.3em}$2{:}\ (1.2,\langle0.0,0.0,0.0,0.1,0.1\rangle)$ \\ $3{:}\ (0.3,\langle0.0,0.0,0.0,0.2,0.0\rangle)$ \vspace{0.2em}} & $0.09$ \\ \hline
\end{tabular}
}

\end{table}

An example execution of the reward machine (RM) constructed from the OCATA in Fig.~\ref{fig:ocata} for the STL specification, is shown in Table~\ref{tab:rm_execution}. We use $r \coloneqq 0.1$ as the reward scaling factor, $x_{\mathit{f}} \coloneqq 4$ as the target position, and
\begin{align}
h(z) \coloneqq \max\{0,\min\{1, 0.5+z\}\},
\; \; z \coloneqq 1-\lvert x-x_{\mathit{f}}\rvert,
\end{align}
as the CDF that determines perturbation probabilities. This choice yields transition probability \(0.5\) at the boundaries of the target region \([3,5]\). The probability varies linearly in the near-boundary bands \([2.5,3.5]\) and \([4.5,5.5]\), and saturates to \(1\) or \(0\) outside these bands.

At $t=1$, the $\epsilon$-action commits to satisfying the ``globally'' component of the formula, moving all probability mass from $l_0$ to $l_1$, regardless of the observation. At $t=2$, the observed position is $x=3.1$, i.e., at distance $0.1$ from the boundary. This yields a probability $0.6$ of remaining in $l_1$. The remaining mass ($0.4$) flows to the conjunction, corresponding to leaving the region under near-boundary behavior due to perturbations. This $0.4$ splits evenly: $0.2$ returns to $l_1$ and $0.2$ transitions to $l_2$ in a newly created memory entry in which the clock is reset. As a result, the first entry has total mass $0.6+0.2=0.8$ in $l_1$ and the second has mass $0.2$ in $l_2$.

At $t=3$, the observation lies exactly on the boundary; thus, the induced transition probabilities are $0.5$. For the first entry, $0.8 \cdot 0.5 = 0.4$ remains in $l_1$, while the other $0.4$ flows to the conjunction; half of that ($0.2$) returns to $l_1$ and the other half ($0.2$) moves to a newly created third entry with a freshly reset clock. For the second entry, because the clock valuation $0.9$ is below the constraint $1.0$, no mass transitions to the rejecting sink. Instead, $0.2 \cdot 0.5 = 0.1$ remains in $l_2$, and the remaining mass transitions to the accepting sink $l_4$ according to the boundary-induced probability $0.5$.

Finally, at $t=4$, the observed position lies well inside the target region; therefore, the transition probabilities saturate to $1$ and $0$. In the first entry, this results in $0.6$ mass remaining in $l_1$. In the second entry, $0.1$ transitions to the rejecting sink $l_4$ since the clock valuation $1.2$ exceeds the threshold $1$, even though the observation is within the target region. In the third entry, all mass ($0.2$) transitions to the accepting sink $l_3$ since the clock valuation $0.3$ is below $1$. Rewards are computed at each step by summing the probabilities in the accepting locations $l_1$ and $l_3$ and multiplying by $r \coloneqq 0.1$.

\subsection{Product Construction for Markovian Rewards}

An RM constructed for STL receives an observation from the SMDP, and an $\epsilon$-action from the RL agent, and outputs a memory list and a scalar reward. The memory list can be used to construct an augmented state, which makes the provided rewards Markovian. This can be formalized by constructing a product as follows:

\begin{definition}
A product MDP is a tuple $\mathcal{M}^\times{=}(S^\times,s_0^\times,\allowbreak A^\times, P^\times, R^\times)$ composed of an MDP $\mathcal{M} {=} (S, s_0,\allowbreak A, P)$ and an STL-RM $\mathcal{R}_\varphi{=}(Q, q_0, \Delta, R)$ such that $S^\times{=}S {\times} \mathbb{R}_{\geq 0} {\times}  Q$ is the set of product states; $s_0^\times{=}\langle s_0,0,q_0\rangle$ is the initial product state; $A^\times{=}A{\times} \mathcal{E}$ is the set of product actions; $P^\times$ is the probabilistic product transition function where $P^\times(\langle s',\tau', q'\rangle \mid \langle s,\tau, q\rangle, \langle a,\varepsilon\rangle) \coloneqq$
$$
\begin{cases}
P(s', \tau'\mid s,a) & \textnormal{ if } q'=\Delta(q,s,\tau,\varepsilon) \\
0 & \textnormal{ otherwise }
\end{cases};
$$
and $R^\times:S^\times\mapsto[0,1]$ is the product reward function where $R^\times(\langle s,\tau,q\rangle) \coloneqq R(q)$.
\end{definition}

The product MDPs constructed in this way are standard discrete-time MDPs and can be readily used with off-the-shelf RL tools. The effect of time durations on STL satisfaction is captured by the memory list, which is part of the product state. The main challenge is the unbounded growth of the memory lists. We believe this is fundamentally difficult to avoid because the observed durations are real-valued, which, without additional assumptions, could require arbitrarily many entries in any approach.
Nevertheless, our approach is substantially more efficient than augmenting the state space with the entire sequence of visited states, as is often required for general unbounded formulas in existing works. In effect, we store only the information needed to determine STL satisfaction from the observations. The efficiency can be further improved by pruning entries that have fully transitioned into sink locations after accounting for their remaining cumulative future reward. Another possible improvement is to store intervals rather than single clock valuations in the entries, especially when the sampling rate is high relative to the time intervals in the STL specifications.

For simplicity, we assume the memory list has a fixed capacity of $N$ entries. Once the list is full, any newly created entry (and its associated probability mass) is discarded, which is equivalent to sending that mass to a rejecting sink state. We now formalize that achieving the maximal expected return of $1$ is only possible under a policy that satisfies the STL specification with probability $1$.

\begin{proposition}
Given an SMDP $\mathcal{M}$ and an STL specification $\varphi$, suppose there exists a product policy $\pi^\times$ that achieves the maximum expected return $1$ in the product MDP $\mathcal{M}^\times=\mathcal{M}\times\mathcal{R}^\times$, formed from $\mathcal{M}$ and the STL-RM $\mathcal{R}^\times$ constructed from $\varphi$. Then, the induced policy $\pi$ in $\mathcal{M}$ satisfies $\varphi$ with probability $1$; i.e.,
\begin{align}
    \mathbb{E}_{\sigma^\times \sim \mathcal{M}^\times_{\pi^\times}} \!\bigl[G^\times(\sigma^\times)\bigr] = 1 \;\Longrightarrow\; \textnormal{Pr}_{\sigma \sim \mathcal{M}_\pi}(\sigma \models \varphi)=1,
\end{align}
where $\sigma^\times \coloneqq \langle s_0,\tau_0,q_0\rangle \langle s_1,\tau_1,q_1\rangle \dots$ denotes a product path sampled from the product Markov chain $\mathcal{M}^\times_{\pi^\times}$ under $\pi^\times$, and $G^\times(\sigma^\times)$ is its associated return. Likewise, $\sigma \coloneqq (s_0,\tau_0)(s_1,\tau_1)\dots$ denotes the timed path sampled from $\mathcal{M}_\pi$ under $\pi$.
\end{proposition}

This follows from the one-to-one correspondence between product paths and timed paths due to the fact that the sequence $q_0,q_1,\dots$ of additional memory-list components along a product path can be uniquely determined by the update function $\Delta$. Now, the return of a product path $\sigma^\times$ is $1$ iff a full reward of $r$ is obtained at each time step (with discount factor $\gamma=1-r$). We note that, under the bounded-memory assumption, newly created entries and their associated probabilities are discarded when the list is full, resulting in to permanent reward loss. Therefore, a return of $1$ requires that no entries are ever discarded and that, at each time step, the total probability mass $1$ is distributed exclusively among accepting locations of the entries in the memory list. This immediately implies that each successor of any conjunctive branching resides in an accepting location, thereby ensuring the B\"uchi condition in each computational thread and implying that the corresponding timed path $\sigma$ satisfies the STL formula $\varphi$.

\begin{figure*}[ht]
    \centering
    \includegraphics[width=\textwidth]{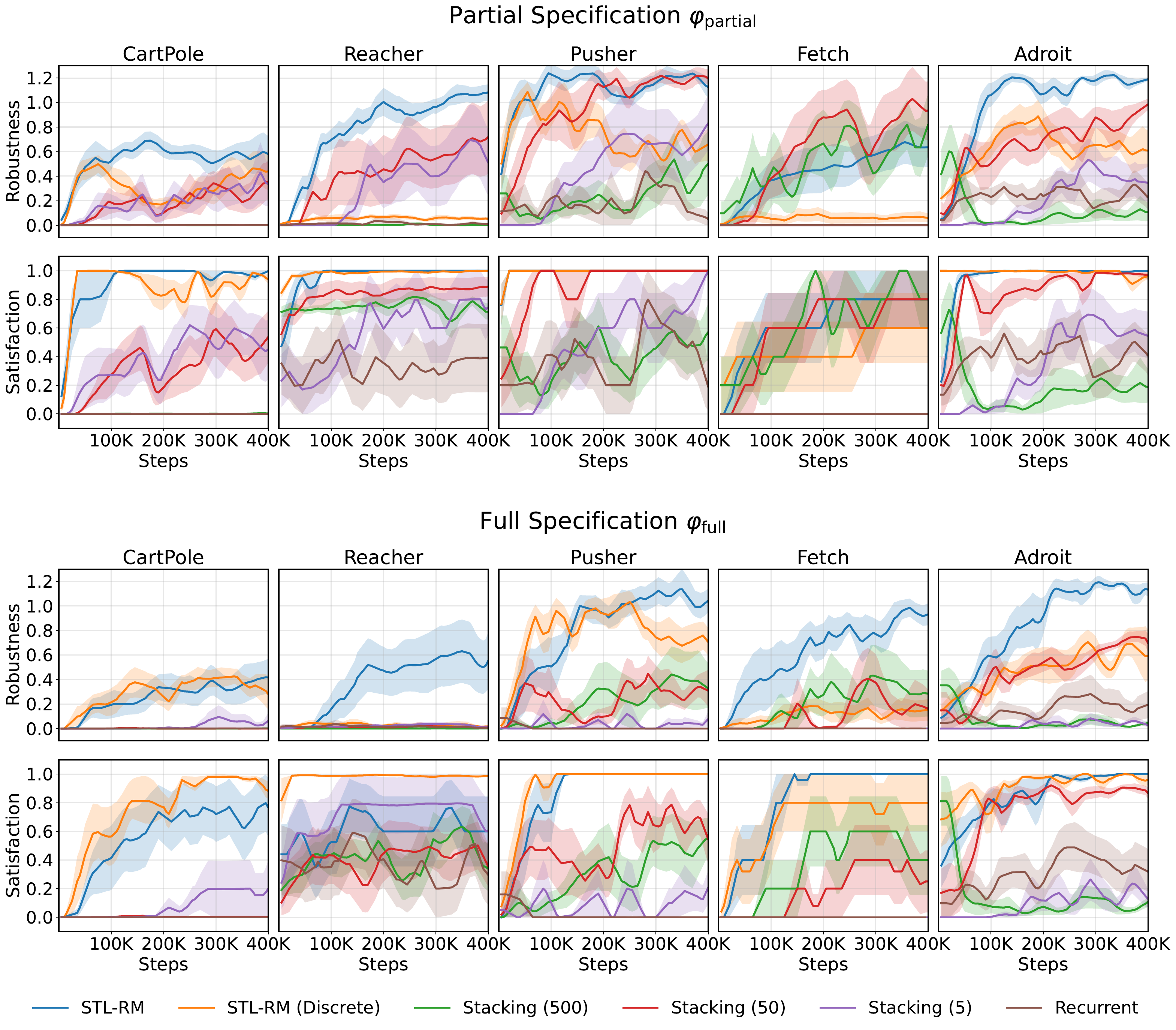}
    \caption{Learning curves of our approach and baselines across all environments. Each method is evaluated every 5K environment steps over 100 episodes using the robustness score and satisfaction rate. Curves and shaded regions show the mean and half the standard deviation, respectively, over five seeds. For visualization, curves are smoothed with a max filter followed by a moving-average filter with window size 5.}
    \label{fig:learning_curves}
\end{figure*}

\section{Experiments}\label{sec:exp}

In this section, we evaluate STL-RM across a diverse set of control environments and compare it against standard history-based RL baselines. We first introduce the STL specifications considered throughout the experiments. We then describe the evaluation environments, baselines, and implementation details, and finally present and discuss the experimental results. Our code is available at \url{https://github.com/alperkamil/stlrm}.

\subsection{STL Tasks}

We consider STL specifications that combine representative temporal requirements. Specifically, we define
\begin{align}
\varphi_{\text{sequencing}}
&\coloneqq F(a_1 \wedge F a_2), \\
\varphi_{\text{stability}}
&\coloneqq
G\left((a_1 \vee a_2)
\implies F_{[0,T]}(b_1 \wedge b_2)\right), \\
\varphi_{\text{safety}}
&\coloneqq G\neg(c_1 \vee c_2), \\
\varphi_{\text{partial}}
&\coloneqq
\varphi_{\text{sequencing}}
\wedge
\varphi_{\text{safety}}, \\
\varphi_{\text{full}}
&\coloneqq
\varphi_{\text{sequencing}}
\wedge
\varphi_{\text{stability}}
\wedge
\varphi_{\text{safety}}.
\end{align}

The three subformulas encode the following temporal behaviors:
\begin{itemize}[leftmargin=*]
    \item \textsc{Sequencing:}
    $\varphi_{\text{sequencing}}$ requires the agent to first visit the target region represented by $a_1$ and subsequently visit the target region represented by $a_2$.

    \item \textsc{Stability:}
    $\varphi_{\text{stability}}$ requires the agent to return to a designated central region, represented by $b_1 \wedge b_2$, within $T$ time steps after visiting either target region $a_1$ or $a_2$.

    \item \textsc{Safety:}
    $\varphi_{\text{safety}}$ requires the agent to avoid the unsafe regions represented by $c_1$ and $c_2$ at all times.
\end{itemize}

Across all environments, the predicates are defined over a one-dimensional signal $x$ extracted from the environment state. The sequencing requirement asks the agent to first reach the positive target region $(3,\infty)$ and subsequently reach the negative target region $(-\infty,-3)$. After visiting either target region, the stability requirement requires the agent to return to the central region $(-2,2)$ within $T=30$ time steps. In addition, the safety requirement constrains the agent to remain within the safe interval $(-6,6)$ throughout the episode, thereby avoiding the unsafe regions $(6,\infty)$ and $(-\infty,-6)$.

The corresponding predicate functions are
\begin{align}
\mu_{a_1}(s) &\coloneqq x-3, &
\mu_{a_2}(s) &\coloneqq -(x+3), \\
\mu_{b_1}(s) &\coloneqq x-2, &
\mu_{b_2}(s) &\coloneqq -(x+2), \\
\mu_{c_1}(s) &\coloneqq x-6, &
\mu_{c_2}(s) &\coloneqq -(x+6),
\end{align}
where each predicate is satisfied whenever its corresponding predicate function is nonnegative.

We evaluate two variants of the STL task, $\varphi_{\text{partial}}$ and $\varphi_{\text{full}}$, which differ in the temporal information required for successful control. The partial specification, $\varphi_{\text{partial}}$, requires ordered visitation of the target regions while enforcing safety. Since it contains only unbounded temporal operators, it abstracts away the precise timing of events and primarily requires the agent to retain their logical progression. Consequently, the task can be represented by a small automaton consisting of, for example, an initial mode, a mode indicating that $a_1$ has been reached, a mode indicating that the sequence has been completed by subsequently reaching $a_2$, and a rejecting mode entered upon reaching either unsafe region $c_1$ or $c_2$, thereby encoding progress through the specification. The functionality of these modes may potentially be captured from a relatively short history of past observations.

In contrast, the full specification, $\varphi_{\text{full}}$, additionally incorporates the bounded-time stability requirement. Whenever either $a_1$ or $a_2$ is visited, the agent must not only remember that the event occurred but also retain its temporal context to ensure that the designated central region is reached within $T$ time steps. Satisfying the full specification therefore requires memory that captures both the logical progression of events and their timing. This explicit real-time constraint is substantially more difficult to directly infer from a history of observations, motivating the structured memory mechanism provided by our STL-RM approach.

\subsection{Environments}

We evaluate STL-RM on five benchmark control environments from Gymnasium~\cite{towers2024gymnasium}. These environments span a broad range of control settings, from low-dimensional discrete control to high-dimensional continuous robotic manipulation:
\begin{itemize}[leftmargin=*]
\item \textsc{CartPole:}
A classical control task with a discrete action space. We use the horizontal position of the cart as the signal for evaluating the STL specification.

\item \textsc{Reacher:}
A continuous-control robotic arm environment. We use the horizontal position of the arm's fingertip as the STL signal.

\item \textsc{Pusher:}
A higher-dimensional continuous-control manipulation task in which a robotic arm interacts with an object. We use the horizontal position of the arm's fingertip as the STL signal.

\item \textsc{Fetch:}
A continuous-control robotic manipulation environment. As in Reacher and Pusher, we use the horizontal position of the robot's end effector as the STL signal.

\item \textsc{Adroit:}
A high-dimensional dexterous manipulation environment. We use the horizontal position of the robotic hand as the STL signal.
\end{itemize}

All environments have continuous observation spaces, whereas their action spaces are continuous except for CartPole, which has a discrete action space. Since the raw position ranges differ across environments, we normalize the selected signal so that its initial mean is zero and apply an environment-specific scaling factor. This transformation places the relevant signal values within a reachable range while allowing the same STL specification to be applied consistently across all environments.

Each episode has a maximum horizon of $500$ time steps. To ensure that learning is driven exclusively by the STL objective rather than by environment-specific objectives, we remove the native reward functions of the environments. We also disable their default early-termination conditions, except for termination resulting from safety violations.

\subsection{Baselines and Implementation}

We compare STL-RM against two standard approaches for incorporating temporal information into RL: \emph{observation stacking} and \emph{recurrent policies}. These baselines provide the policy with access to historical information without explicitly constructing a symbolic representation of the temporal specification.

\begin{itemize}[leftmargin=*]
\item \textsc{STL-RM} (Ours):
We implement STL-RM in Python. For the specification defined above, we construct standard B\"uchi automata for $\varphi_{\text{sequencing}}$ and $\varphi_{\text{safety}}$. The bounded-time stability specification $\varphi_{\text{stability}}$ is represented using an OCATA together with the memory mechanism introduced in our method. The memory-list capacity is set to $N=50$. For transitions leading to accepting and rejecting states (i.e., non-accepting sink states), we use the clipped functions $h_{\text{accept}}(z)=\max\{0,\min\{1,z\}\}$ and $h_{\text{reject}}(z)=\max\{0,\min\{1,1+z\}\}$, respectively, where $z$ denotes the distance between the transition boundary and the current signal value. For simplicity, we set the discount factors to the PPO default values, $\gamma=\gamma'=0.99$.

\item \textsc{STL-RM (Discrete)} (Ours):
This ablation is obtained by discretizing the transitions in STL-RM. Specifically, the CDF is replaced by the indicator function $h(z)=\mathbbm{1}_{z\geq 0}$, which evaluates to $1$ when $z\geq0$ and to $0$ otherwise. This variant allows us to isolate the effect of the continuous robustness-aware transition construction used in STL-RM.

\item \textsc{Stacking}:
This baseline explicitly represents temporal history by concatenating a fixed number of previously observed signal values with the current environment observation. We consider stack sizes of $5$, $50$, and $500$, corresponding to short, intermediate, and full-episode histories, respectively. These configurations allow us to examine how the amount of explicitly available history of observations affects the performance of a feedforward policy.

\item \textsc{Recurrent}:
This baseline uses a long short-term memory (LSTM) network~\cite{hochreiter1997long} to encode observation history. Unlike observation stacking, which explicitly retains a fixed history window, the LSTM learns a latent representation of temporally relevant information. We use a single LSTM layer with $256$ hidden units.
\end{itemize}

To the best of our knowledge, existing RL methods for STL do not directly support the class of general formulas with arbitrarily nested temporal operators considered. We therefore use these two history-based approaches as generic baselines that work for entire STL for comparison. We evaluate whether the explicit STL-based memory and reward-machine structure of \textsc{STL-RM} provide an advantage over learned representations of observation history via robustness scores.

All methods use Proximal Policy Optimization (PPO)~\cite{schulman2017proximal}, implemented with Stable-Baselines3~\cite{raffin2021stable}, as the underlying RL algorithm. The actor and critic networks consist of two fully connected hidden layers with $256$ units per layer. The recurrent baseline additionally includes the $256$-unit LSTM layer described above. All remaining PPO hyperparameters are set to their Stable-Baselines3 default values.

For both \textsc{Stacking} and \textsc{Recurrent}, we use dense-time online STL robustness computed with RTAMT~\cite{yamaguchi2024rtamt} as the reward signal at each training step. This provides the baselines with a dense STL-based learning signal while leaving the policy responsible for representing the temporal history required to satisfy the specification.

During evaluation, we use the same metrics for all methods, including \textsc{STL-RM}. Specifically, we report the final average STL robustness score over 100 evaluation trajectories, together with the corresponding STL satisfaction rate, enabling a direct comparison of the methods based on their ability to satisfy the complete temporal specification. For the STL specifications considered in our experiments, the maximum achievable robustness score is $1.5$, whereas the minimum is unbounded below. To prevent large negative values from disproportionately affecting the reported averages, we clip negative robustness scores to zero.

\subsection{Results}

Fig.~\ref{fig:learning_curves} presents the learning curves, obtained over 400 K steps (K=1,000), of all methods across the five environments. For the partial specification $\varphi_{\text{partial}}$, both \textsc{STL-RM} and \textsc{STL-RM (Discrete)} rapidly converge to policies with a satisfaction rate of $1$. The two variants of our approach outperform the baselines in terms of satisfaction rate across all environments except \textsc{Fetch}, where the baselines achieve competitive performance. Moreover, \textsc{STL-RM} generally achieves higher robustness scores, indicating that its robustness-aware transition construction encourages policies that satisfy the specification with larger margins.

The baselines exhibit a substantial performance degradation when moving from the partial specification $\varphi_{\text{partial}}$ to the full specification $\varphi_{\text{full}}$. This result highlights the difficulty of learning bounded-time requirements from observation history using STL robustness alone as the reward signal. In contrast, \textsc{STL-RM} and \textsc{STL-RM (Discrete)} converge to optimal or near-optimal policies in terms of satisfaction rate, while \textsc{STL-RM} achieves higher robustness scores than the baselines. The performance of both STL-RM variants improves in \textsc{Fetch} under the full specification, suggesting that the additional stability requirement can provide a useful learning signal in this environment.

The difference in satisfaction rate between \textsc{STL-RM} and \textsc{STL-RM (Discrete)} is generally small across environments. However, \textsc{STL-RM} typically achieves higher robustness scores. This observation suggests that although discretized automaton transitions may be sufficient to learn satisfying policies, incorporating continuous robustness information into the reward-machine transitions through probabilistic transitions encourages trajectories that satisfy the specification more robustly.

\section{Conclusion}\label{sec:conclusion}

This paper advances RL from general STL specifications by addressing the central challenge posed by their history-dependent satisfaction semantics. By compiling event-based STL formulas into OCATAs and subsequently into STL-RMs, our framework provides a compact memory representation and Markovian reward structure compatible with standard RL algorithms. In doing so, STL-RM avoids the computational burden of explicitly augmenting the state with observation histories while retaining the expressive power of STL beyond the restricted fragments considered in prior work. We further incorporate robustness to observation perturbations into the reward construction and establish that policies maximizing the cumulative reward satisfy the specification almost surely. Our empirical results demonstrate that STL-RM achieves higher satisfaction rates and robustness scores than history-based baselines trained directly with STL robustness rewards, indicating that automata-based reward design provides a promising direction for model-free control synthesis from expressive STL specifications.

A fundamental limitation of the proposed approach is its finite memory capacity. Although the STL-RM memory is substantially more compact than explicitly storing past observations, the memory list may require a new entry for each temporally constrained condition that must be tracked. Consequently, the memory capacity can become a bottleneck for specifications that generate many simultaneous temporal obligations. An important direction for future work is therefore to develop more compact memory representations, for example by integrating interval-based representations that aggregate multiple related memory entries rather than storing them individually.

Another advantage of STL-RM is that its reward-machine formulation makes a broad range of existing reward-machine extensions readily applicable. For example, the framework could be extended with counterfactual reasoning techniques~\cite{icarte2022reward}, as well as generalized to settings involving lexicographic objectives~\cite{bozkurt2021lexicographic} and stochastic games~\cite{bozkurt2021games,bozkurt2021secure,bozkurt2024learning}. These directions provide opportunities to improve learning efficiency and extend STL-based policy synthesis to richer multi-objective and multi-agent control environments.

\vspace{-1em}
\section*{References}
\vspace{-2.5em}
\bibliographystyle{IEEEtran}
\bibliography{references}

\vspace{-2em}
\begin{IEEEbiography}[{\includegraphics[width=1in,height=1.25in,clip,keepaspectratio]{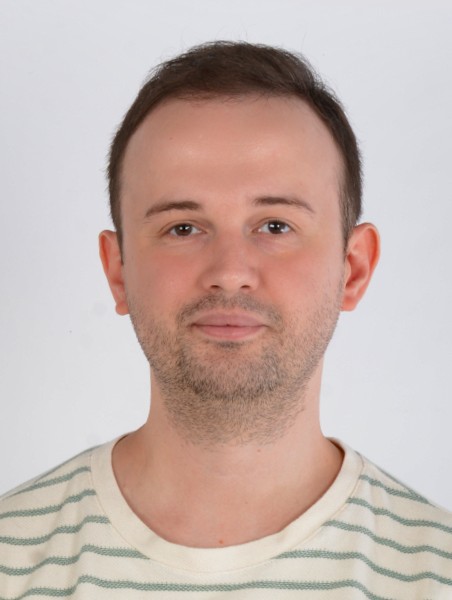}}]{Alper Kamil Bozkurt} received the B.S. and M.S. degrees in computer engineering from Bogazici University, Istanbul, Turkey, in 2015 and 2018, respectively. He obtained a Ph.D. degree in computer science from Duke University, Durham, VA, USA. He is currently a postdoctoral associate at Virginia Commonwealth University, Richmond, VA, USA. Previously, he was a postdoctoral associate at University of Maryland, College Park, MD, USA. His research interests lie at the intersection of machine learning, control theory, and formal methods. In particular, he focuses on developing learning-based algorithms that synthesize provably safe and reliable controllers for robotics and cyber-physical systems.
\end{IEEEbiography}
\vspace{-2em}
\begin{IEEEbiography}[{\includegraphics[width=1in,height=1.25in,clip,keepaspectratio]{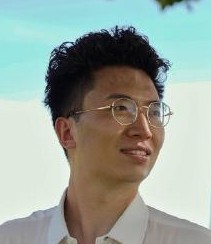}}]{Shangtong Zhang} is an Assistant Professor in the Department of Computer Science at the University of Virginia, Charlottesville, VA, USA, directing the Sequential Intelligence Lab (SIL). His research focuses on both theoretical and empirical aspects of reinforcement learning, resulting in multiple scholarly articles in major AI venues, e.g., JMLR, NeurIPS, ICML, and ICLR. He also regularly serves as Area Chair in major AI venues, e.g., NeurIPS, ICML, ICLR, Senior Area Chair in RL Conference, Action Editor in TMLR, and panelists and reviewers for major federal (e.g., NSF) and international (e.g., Schmidt Sciences) funding agencies. He and his research are recognized by multiple awards and honors, including best paper awards at ICML workshop and AAMAS, NSF CAREER Award, AAAI New Faculty Highlights, Google Research Award, Cisco Faculty Research Award, Nvidia Academic Grant, Rising Star in AI, NeurIPS Top Area Chair, and IFAAMAS Victor Lesser Dissertation Award (runner-up). He obtained his DPhil at the University of Oxford, Oxford, UK, MSc at the University of Alberta, Edmonton, AB, Canada, and BSc at Fudan University, Shanghai, China.
\end{IEEEbiography}
\vspace{-2em}
\begin{IEEEbiography}[{\includegraphics[width=1in,height=1.25in,clip,keepaspectratio]{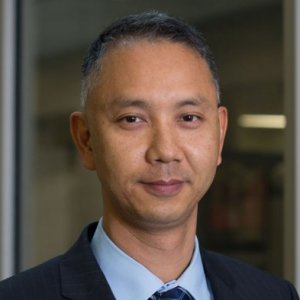}}]{Yuichi Motai} received the B.Eng. degree in instrumentation engineering from Keio University, Tokyo, Japan, in 1991, the M.Eng. degree in applied systems science from Kyoto University, Kyoto, Japan, in 1993, and the Ph.D. degree in electrical and computer engineering from Purdue University, West Lafayette, IN, USA, in 2002. He is currently an Associate Professor of Electrical and Computer Engineering at Virginia Commonwealth University, Richmond, VA, USA. His research interests include the broad area of sensory intelligence, particularly in data analytics, pattern recognition, computer vision, and sensory-based robotics.
\end{IEEEbiography}

\end{document}